\documentclass[10pt,twocolumn,letterpaper,conference]{IEEEtran}

\usepackage{times}
\usepackage{epsfig}
\usepackage{graphicx}
\usepackage{amsmath}
\usepackage{caption}
\usepackage{subcaption}
\usepackage{amssymb}
\usepackage{algorithm2e}
\usepackage{changepage}
\usepackage{multirow}
\usepackage[table,xcdraw]{xcolor}
\usepackage{rotating}
\graphicspath{ {imgs/} }
\usepackage{comment}
\usepackage{balance}

\begin{document}

\title{\LARGE Harnessing Unlabeled Data to Improve Generalization of Biometric Gender and Age Classifiers}


 \author{\IEEEauthorblockN{Aakash Varma Nadimpalli$^{1}$, Narsi Reddy$^{2}$, Sreeraj Ramachandran$^{1}$ and Ajita Rattani$^{1}$}
\IEEEauthorblockA{\textit{$^{1}$School of Computing, Wichita State University, Wichita, USA}\\ 
\textit{$^{2}$School of Computing, University of Missouri, Kansas City, USA}\\ 
Email: \{axnadimpalli,sxramachandran2\}@shockers.wichita.edu}, ajita.rattani@wichita.edu
}

\maketitle

\begin{abstract}
With significant advances in deep learning, many computer vision applications have reached the inflection point. However, these deep learning models need large amount of labeled data for model training and optimum parameter estimation. Limited labeled data for model training results in over-fitting and impacts their generalization performance. However, the collection and annotation of large amount of data is a very time consuming and expensive operation. Further, due to privacy and security concerns, the large amount of labeled data could not be collected for certain applications such as those involving medical field. Self-training, Co-training, and Self-ensemble methods are three types of semi-supervised learning methods that can be used to exploit unlabeled data.
In this paper, we propose self-ensemble based deep learning model that along with limited labeled data, harness unlabeled data for improving the generalization performance. 
We evaluated the proposed self-ensemble based deep-learning model for soft-biometric gender and age classification. 
Experimental evaluation on CelebA and VISOB datasets suggest gender classification accuracy of \textbf{94.46\%} and \textbf{81.00\%}, respectively, using only \textbf{1000} labeled samples and \textbf{remaining 199k samples as} unlabeled samples for CelebA dataset and similarly,\textbf{1000} labeled samples with \textbf{remaining 107k samples as} unlabeled samples for VISOB dataset. Comparative evaluation suggest that there is $5.74\%$ and $8.47\%$ improvement in the accuracy of the self-ensemble model when compared with supervised model trained on the entire CelebA and VISOB dataset, respectively.
We also evaluated the proposed learning method for age-group prediction on Adience dataset and it outperformed the baseline supervised deep-learning learning model with a better exact accuracy of \textbf{55.55 $\pm$ 4.28}  which is \textbf{3.92\%} more than the baseline.
\end{abstract}

\IEEEoverridecommandlockouts

\IEEEpeerreviewmaketitle


\section{Introduction}
\label{sec:intro}
AI and computer vision has reached inflection point with advances in deep learning. However, these high accuracy rates are obtained at the cost of large number of parameters involved which in turn require large scale labeled (\textbf{annotated}) datasets for network training and learning optimal set of parameters.

The availability of large-scale annotated datasets may be limited due to the requirement of manual annotation by the human expert which is a cumbersome and time consuming operation. 
Further, due to the nature of field, large datasets may not be publicly available, for instance, in the medical field, due to the privacy and security concerns. Training using limited datasets impact the generalization ability of the deep learning models due to overfitting that arise due to  small number of data points. 

A number of studies have explored the automatic extraction of demographic attributes such as gender, age, ethnicity, etc. of an individual, known as soft-biometrics~\cite{jain2004soft}. These attributes have been deduced from biometric data such as facial images, voice, gait, and hand or body images. Automated soft-biometric prediction has drawn significant interest in numerous applications such as surveillance, human–computer interaction, anonymous customized advertisement system, image retrieval system, continuous user authentication, subject re-identification, and in fusion with primary biometric modalities for performance enhancement~\cite{reid2013soft}.

                                                                                    \begin{figure}[t!]
                                                                                    \centering
                                                                                    \includegraphics[scale=0.22]{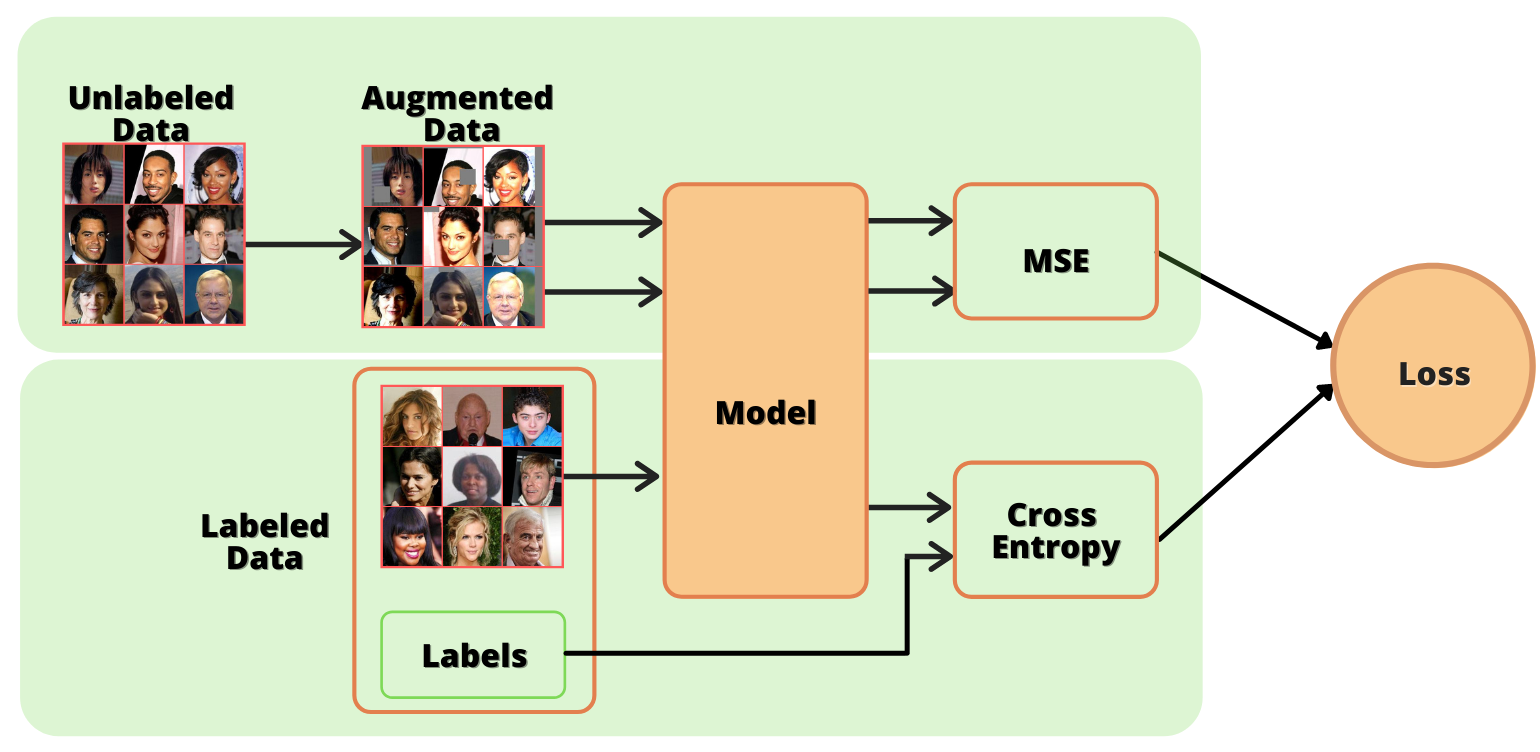}
                                                                                    \caption{Schema of the proposed self-ensemble model with labeled and unlabeled data, respectively. The joint-loss function which is the combination of the MSE (Mean Squared Error) and Cross-Entropy loss is used for model training using both labeled and unlabeled data.}
                                                                                    \label{img:pi_model}
                                                                                    \end{figure}

Studies have been conducted to improve the prediction accuracy of soft biometrics in different biometric traits~\cite{dantcheva2016else} using deep learning models. For face biometrics, studies have been conducted for deducing gender~\cite{levi2015age, zhang2017age, narang2016gender}, age~\cite{levi2015age, zhang2017age, zhang2018fine}, and ethnicity~\cite{narang2016gender, masood2018prediction} from facial images. 
Soft biometrics such as gender~\cite{thomas2007learning, lagree2011predicting, rattani2018convolutional, Bobeldyk2019Pixels, tapia2018sex}, ethnicity~\cite{lagree2011predicting, Bobeldyk2019Pixels, mohammad2018convolutional}, age~\cite{sgroi2013prediction, rattani2017convolutional}, and eye color~\cite{bobeldyk2018predicting, Bobeldyk2019Pixels} have been predicted from periocular and iris in visible and near infrared spectrum.

Studies in biometric recognition have shown steady improvement in matching performance by combining soft biometric attributes with primary biometrics~\cite{jain2009facial, gonzalez2018facial, jain2004soft}. In \cite{jain2004soft}, showed around $5\%$ improvement in identification system and around $4\%$ improvement in verification system by combining primary biometrics with predicted soft biometric traits such as ethnicity, gender, and height. 

Recent advancements in soft biometric predictions are mainly based on deep learning for feature extraction and classification such as in ~\cite{levi2015age, Bobeldyk2019Pixels, rattani2018convolutional}. These deep learning methods require a large amount of training data to obtain high generalization accuracy rates. However, acquisition of large biometric datasets along with soft-biometric attributes such as gender, race and age, may invokes privacy and security issues. 

One way to overcome this problem is by using semi-supervised learning~\cite{zhu2005semi}. Semi-supervised learning entails joint use of labeled and unlabeled data in improving the generalization performance of the classifier. 

In this paper, we proposed a self-ensemble based semi-supervised learning method which improve the generalization ability of the deep learning model trained on limited labeled data and large amount of unlabeled data.
\textbf{The proposed method develops a joint loss function that uses categorical cross-entropy for the labeled samples while reducing the distance between the model outputs of two different perturbations on the unlabeled data}. Figure~\ref{img:pi_model} show the schema of the proposed self-ensemble model. 
Many of the semi-supervised methods propose to use noise to induce perturbations~\cite{rasmus2015semi, laine2016temporal}. However, using noise might affect the features which are used in predicting soft biometric attributes to be deduced from biometric traits. Therefore, we proposed to use data augmentation to generate different perturbations from unlabeled data. 

The contributions of this paper are two-fold:
\begin{itemize}
    \item The proposal of \textbf{self-ensemble based deep learning model} which harness unlabeled data to improve the generalization ability of the soft-biometric prediction. 
    \item Experimental evaluations are conducted for \textbf{gender prediction} on eye region images from VISOB dataset and face images from CelebA dataset. We have also evaluated \textbf{age-group prediction} on face images from Adience dataset. We show the deep learning models can be trained using as low as $100$ labeled samples using the proposed semi-supervised learning method.
\end{itemize}


The rest of the paper is organized as follows: A brief explanation of semi-supervised learning with recent deep learning based advancements is provided in section-\ref{sec:semi_sup}. The proposed self-ensemble method is discussed in section-\ref{sec:proposed_method}. Experimental setup and datasets are presented in section-\ref{sec:exp_setup}. Results are discussed in section-\ref{sec:results} and paper is concluded in section-\ref{sec:conclusion}.


\section{Prior Work on Semi-Supervised Learning}
\label{sec:semi_sup}

Semi-supervised learning is a learning technique, which make joint use of labeled and unlabeled data for training a classifier, have a long history in machine learning research~\cite{zhu2005semi}. Specifically, the model trained on labeled data is used to generate pseudo-label for the unlabeled data. The pseudo-labeled unlabeled data is used for re-training the classifier for enhanced accuracy. For misclassification in psuedo-labeled data, the deep learning models have proven to be good at dealing with noisy labels~\cite{Reed2015NoisyLabels}. 


These semi-supervised learning methods can be broadly divided into three types as follows:

\textbf{(1) Self-training:} In self-training based semi-supervised learning, a single model trained on a small labeled dataset is used to generate pseudo-labels for unlabeled dataset. The labeled data together with pseudo-labeled unlabeled data is used to re-train the model~\cite{suzuki2008semi, lee2017deep}. Self-training is one of the earliest methods in semi-supervised learning and is used for performance enhancement of the classifier.

\textbf{(2) Co-training:} 
Co-training is a suitable semi-supervised learning approach that needs the dataset to be characterized by two different interpretations of features. 
In co-training, joint use of two-classifiers is used for pseudo-label generation and classifier re-training on unlabeled dataset. One way of performing co-training is one or more models $m_{i =  1 \to n}$ are used to predict proxy (pseudo-labels) labels for another model $m_{j \neq i}$\cite{ruder2018strong}. Another way is by doing a majority voting on the predicted labels from two or more models to generate a proxy label for unlabeled dataset which is used for classifier re-training~\cite{zhou2004democratic}.

\textbf{(3) Self-ensemble:} This training process is similar to the co-training, where instead of using multiple classifiers to predict the pseudo-labels, only a single model with different perturbations of input samples is used. In self-ensemble, the model is trained using both labeled and unlabeled data during the training stage. Rasmus et al.~\cite{rasmus2015semi} proposed to train model robust to noise by reducing the prediction error between noisy and clean input of unlabeled dataset while predicting targets for the labeled dataset in a supervised manner. The noise is induced by adding Gaussian noise to the input image as well as to the output of each layer in the model. 
Laine et al.~\cite{laine2016temporal} applied two different perturbations using Gaussian noise and augmentations to unlabeled samples and used dropout in deep learning model instead of Gaussian noise. 

Our proposed method is based on self-ensemble technique, where instead of using noise as a perturbation in the previous models, we propose to use only data augmentations such as \textit{color jitter, flipping the image horizontally and using random translations}. The proposed method is discussed in more detail in Section-\ref{sec:proposed_method}.

\begin{figure}[t!]
\centering
\includegraphics[scale=0.5]{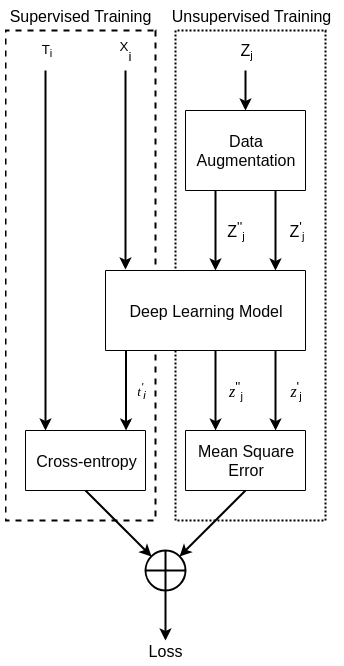}
\caption{Proposed self-ensemble based model with consecutive supervised and unsupervised training components using labeled and unlabeled data, respectively.}
\label{img:pi_model}
\end{figure}

\section{Proposed Method}
\label{sec:proposed_method}
Block diagram of the proposed self-ensemble method is shown in Figure~\ref{img:pi_model}. In the proposed method, the deep learning model is trained consecutively in supervised manner on data $X_i$ with labels $T_i$ and in unsupervised manner by reducing output error between the pseudo-labels $z^{'}_j$ and $z^{''}_j$ of two augmented samples $Z^{'}_j$ and $Z^{''}_j$ generated from the unlabeled data $Z_j$. \emph{This is done so that the supervised part of the training will ensure the model learns to predict the correct target class for the given input while unsupervised part ensures the model learns to produce consistent output}.

\textbf{Data Augmentation for Perturbation:} As we mentioned in Section~\ref{sec:intro}, the proposed method is based on self-ensemble based semi-supervised learning, where two output images were generated by inducing different kind of noise to the input image. However, inducing noise may distort the features in biometric traits, such as ocular and fingerprint images. We therefore propose to use data augmentation. In our experiments, we incorporated two different augmentations: firstly, we applied color jitter by changing the brightness and saturation of the image. Secondly, we randomly flipped the input image horizontally and finally, translation is applied by randomly cropping a region out of the image. This way, from data $Z_j$, we generate two perturbed samples $Z^{'}_j$ and $Z^{''}_j$ to train the deep learning model for the unsupervised part. Using  data augmentation  around 108K unlabeled samples for VISOB dataset and 199k unlabeled samples for CelebA dataset were generated.  

\begin{table}[]
\label{tab:deep_arch}
\caption{Architecture of the proposed self-ensemble based deep learning model that make joint use of labeled and unlabeled data in soft-biometrics prediction.}
{\renewcommand{\arraystretch}{1.1}
\begin{tabular}{lcc}
\hline
\multicolumn{1}{c}{\textbf{Layer}}                                                      & \textbf{Output Layer} & \textbf{Parameters} \\ \hline\hline
Conv 3x3, 32 filters                                                                    & {[}32, 128, 128{]}    & 288                 \\
$\begin{bmatrix}BN, ReLU,\\ CONV \ 3\times3\\32 Filters\end{bmatrix} \times 1$             & {[}32, 128, 128{]}    & 9,280               \\
MaxPool 2x2                                                                             & {[}32, 64, 64{]}      & -                   \\
$\begin{bmatrix}BN, ReLU,\\ CONV \ 3\times3\\64 Filters\end{bmatrix} \times 2$  & {[}64, 64, 64{]}      & 55,488              \\
MaxPool 2x2                                                                             & {[}64, 32, 32{]}      & -                   \\
$\begin{bmatrix}BN, ReLU,\\ CONV \ 3\times3\\128 Filters\end{bmatrix} \times 2$ & {[}128, 32, 32{]}     & 221,568             \\
MaxPool 2x2                                                                             & {[}128, 16, 16{]}     & -                   \\
$\begin{bmatrix}BN, ReLU,\\ CONV \ 3\times3\\128 Filters\end{bmatrix} \times 2$ & {[}128, 16, 16{]}     & 295,424             \\
MaxPool 2x2                                                                             & {[}128, 8, 8{]}       & -                   \\
$\begin{bmatrix}BN, ReLU,\\ CONV \ 3\times3\\128 Filters\end{bmatrix} \times 2$ & {[}128, 8, 8{]}       & 295,424             \\
Global AvgPool 8x8                                                                      & {[}128{]}             & -                   \\
Dense Layer                                                                             & {[}2{]}               & 256                 \\ \hline\hline
\multicolumn{1}{c}{\textbf{Total Parameters}}                                                           & \textbf{}             & \textbf{877,728}    \\ \hline
\end{tabular}
}
\end{table}

\textbf{Proposed Deep Learning Model:} In our experiments, we used a simple sequential convolution neural network (CNN) based on visual geometry group (VGG)~\cite{simonyan2015deep} architecture. Table 1 shows the proposed deep learning model used in our experiments. In all our experiments, we used a single channel (grayscale) image of size $128\times128$ pixels as an input, from which we extract $128$ features by taking the global average of the output of the last convolutional layer. This is followed by the final fully connected layer (dense layer) for predicting the soft-biometric attributes. The proposed deep learning model is relatively small with only $877.7K$ parameters as shown in Table 1.

\textbf{Loss Function:} For a given labeled data $X_i$ with targets $T_i$, the model predicts $T^{'}_i$ as output, then we use cross-entropy loss as shown in equation-\ref{eq:cce} for supervised part of the model training.

\begin{equation}
\label{eq:cce}
    H(T_i, T^{'}_i) = - \sum_{i} T_i \log (T^{'}_i)
\end{equation}

Let $z^{'}_j$ and $z^{''}_j$ be the predictions from the model for the given two perturbed samples, $Z^{'}_j$ and $Z^{''}_j$, for given unlabeled input $Z_j$, then the model is trained in an unsupervised manner by taking mean square error (MSE) between two model predictions as shown in equation~\ref{eq:mse}.

\begin{equation}
\label{eq:mse}
    L(z^{'}_j, z^{''}_j) = \frac{1}{N} \sum \left (  z^{'}_j- z^{''}_j\right )^2
\end{equation}

Finally, the proposed model is trained by joint combination of cross-entropy and mean square error loss functions as shown in equation \ref{eq:loss}. 

\begin{equation}
\label{eq:loss}
    L = H(T_i, T^{'}_i) + \alpha * L(z^{'}_j, z^{''}_j) 
\end{equation}

Where $\alpha$ is coefficient for unsupervised loss (MSE) and in all our experiments, we used $\alpha = 1.0$ based on empirical evidence. 

\section{Experimental Setup}
\label{sec:exp_setup}

\subsection{Datasets}
\label{subsec:dataset}
We evaluated the proposed self-ensemble method on gender and age-group soft biometrics predictions. 
\begin{figure}[]
  \centering
  \label{celebdf_samples}
  
    \subfloat{\includegraphics[width=0.35\textwidth]{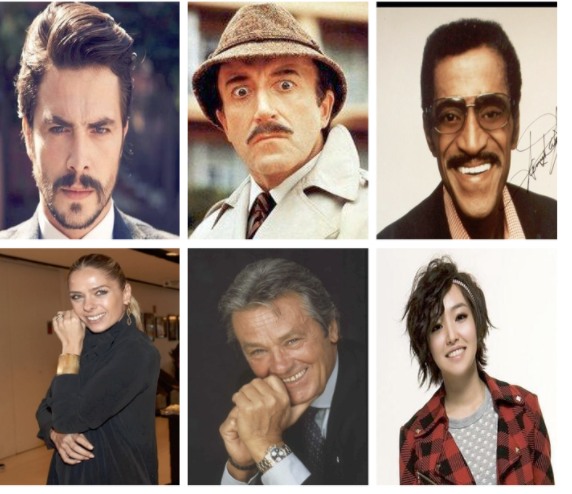}\label{cr_3}}
  \caption{Sample images from CelebA dataset~\cite{liu2015faceattributes} used for evaluation of the self-ensemble based deep learning classifier in gender prediction from facial images.}
\end{figure}

\begin{figure}[]
  \centering
  \label{celebdf_samples}
  \subfloat{\includegraphics[width=0.12\textwidth]{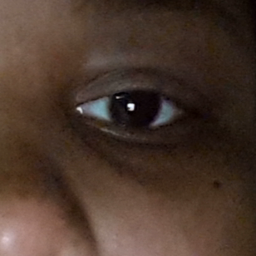}\label{cf_1}}\hspace{0.1cm}
    \subfloat{\includegraphics[width=0.12\textwidth]{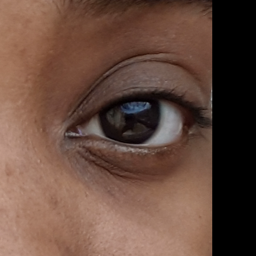}\label{cf_2}}\hspace{0.1cm}
     \subfloat{\includegraphics[width=0.12\textwidth]{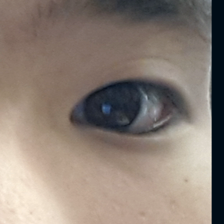}\label{cf_3}}\hspace{0.1cm}
     \subfloat{\includegraphics[width=0.12\textwidth]{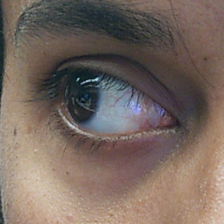}\label{cr_1}}\hspace{0.1cm}
    \subfloat{\includegraphics[width=0.12\textwidth]{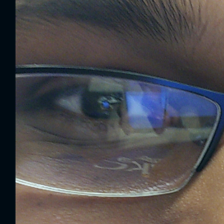}\label{cr_2}}\hspace{0.1cm}
    \subfloat{\includegraphics[width=0.12\textwidth]{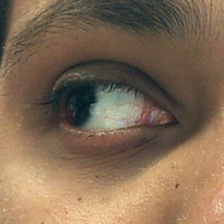}\label{cr_3}}
  \caption{Sample images from VISOB dataset~\cite{VISOB} used for evaluation of Self-ensemble deep-learning based gender classifier from ocular images.}
\end{figure}

\begin{figure}[]
  \centering
  \label{celebdf_samples}
  \subfloat{\includegraphics[width=0.12\textwidth]{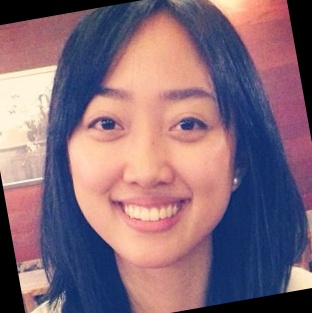}\label{cf_1}}\hspace{0.1cm}
    \subfloat{\includegraphics[width=0.12\textwidth]{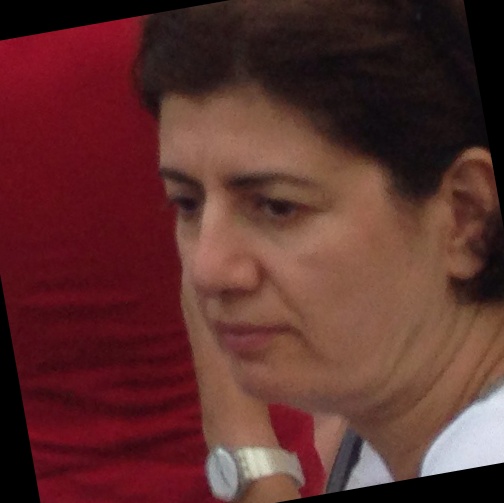}\label{cf_2}}\hspace{0.1cm}
     \subfloat{\includegraphics[width=0.12\textwidth]{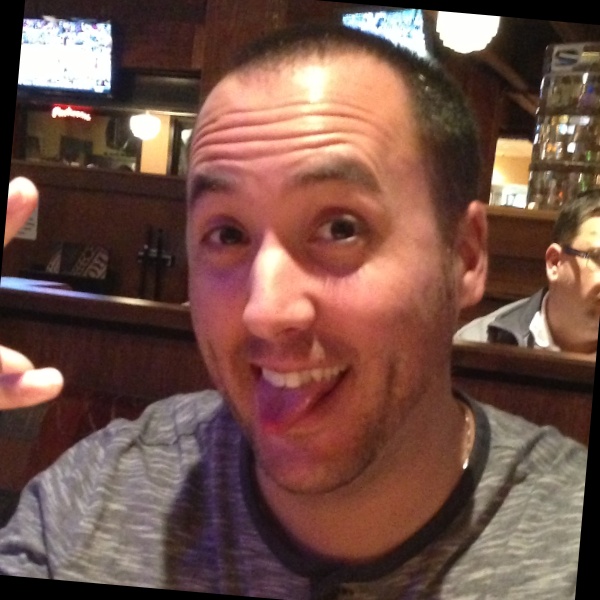}\label{cf_3}}\hspace{0.1cm}
     \subfloat{\includegraphics[width=0.12\textwidth]{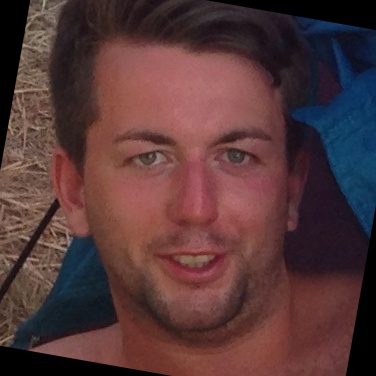}\label{cr_1}}\hspace{0.1cm}
    \subfloat{\includegraphics[width=0.12\textwidth]{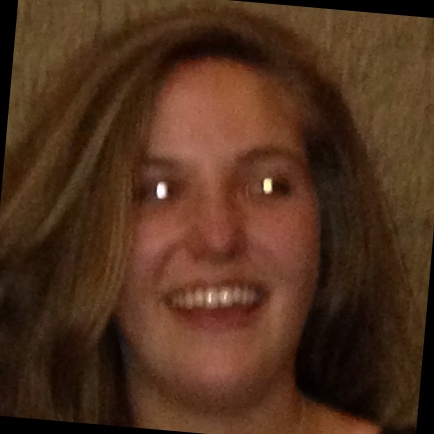}\label{cr_2}}\hspace{0.1cm}
    \subfloat{\includegraphics[width=0.12\textwidth]{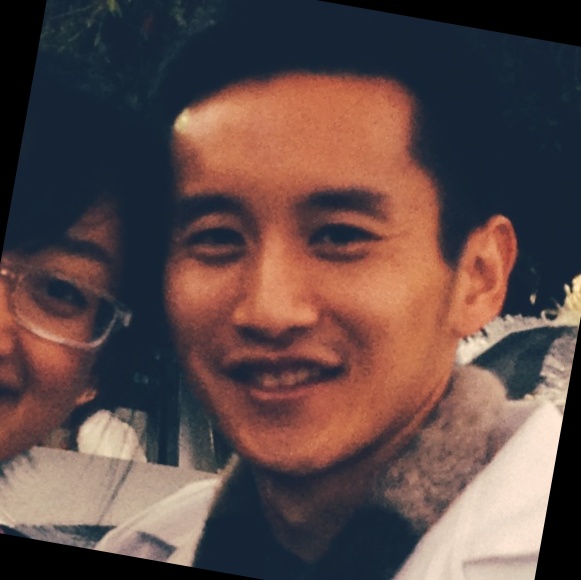}\label{cr_3}}
  \caption{Sample images from Adience dataset~\cite{eidinger2014age} used for evaluation of self-ensemble based age-group classification from facial images.}
\end{figure}

\textbf{Gender Prediction:} For prediction of gender as soft biometric trait, we evaluated the proposed model on eye region in visible light spectrum using VISOB~\cite{VISOB} dataset and also face images in visible light spectrum using CelebA~\cite{liu2015faceattributes} dataset. For gender prediction, we trained the proposed model in subject independent evaluation with very few labeled samples, ranging from $100$ to $1000$ collectively for both genders and a large number of unlabeled samples of about $108$K. 

In the case of VISOB~\cite{VISOB} dataset, we used extended eye region crops used in \cite{rattani2018convolutional} for our evaluation. VISOB dataset consists of eye images from $550$ subjects captured using selfie cameras of three different smartphones (iPhone 5s, Oppo N1 and Samsung Note 4). Eye captures are collected from the subjects in two visits at least $2$ weeks apart in two sessions in each visit in multiple lighting conditions. The dataset is divided into $11$ batches with $50$ subjects each, and we randomly pick $8$ folds for training and validation, and the remaining $3$ folds are used testing. This process is repeated $5$ times, and the average performance is reported. 

CelebA~\cite{liu2015faceattributes} is the publicly available face dataset with more than $200K$ images from $10K$ subjects with 40 soft biometric attributes assigned for each image. In our experiments, we divided the dataset into $70/30$ subject independent split with equal distribution for male and female subjects in both the sets. During training, we randomly picked a small subset, ranging from $100$ to $1000$, of samples as labeled set and rest of the training dataset of about $199$K as the unlabeled set.

We evaluated the performance of gender prediction in terms of male classification rate (MCR$\%$) and female classification rate (FCR$\%$) along with total classification accuracy (ACC$\%$) as reported in \cite{rattani2018convolutional} where MCR is the percentage of male samples correctly predicted as male and similarly for FCR with female samples.

\textbf{Age-Group Prediction:} We evaluated the age-group prediction on Adience~\cite{eidinger2014age} dataset with $8$ age groups ($0-2$, $4-6$, $8-13$, $15-20$, $25-32$, $38-43$, $48-53$, $60-$). The dataset consists of $26K$ face images from $2284$ subjects which are divided into $5$ subject disjoint sets. We followed the cross-validation protocol presented in \cite{eidinger2014age} and reported the performance metrics in terms of exact age-group prediction rate (Exact Acc\%) and one-off age-group prediction rate (1-OFF Acc\%). As the dataset contains unconstrained and unfiltered face samples collected from \textit{flicker.com} albums with large noise induced due to motion blur, resolution, type of capturing device, gaze, and illumination, this rendered the age-group prediction to be a complex task. For this reason, we used all the samples of about $11$K in training set, as both labeled and unlabeled set (50:50 split), to train the proposed self-ensemble based deep learning model.  

\subsection{Experimental Protocol}
\label{subsec:train_protocol}
For both gender and age-group prediction, first we trained the proposed model shown in Table 1, as the baseline with all the training dataset as labeled samples. Once we obtained the best results from our experiments, we used the same global parameters for the rest of our experiments. For gender prediction, we used a batch size of $32$ with a learning rate of $0.001$ and trained the model for $50$ epochs. In the case of age-group prediction, we used a batch size of $64$ with a learning rate of $0.0001$ and trained for $200$ epochs. For both gender and age group predictions, we used Adam~\cite{kingma2017adam} optimizer.

Before applying our proposed data augmentation pipeline, we resized all the images to $144\times144$ pixels. For each batch of the given image, we randomly apply color jitter followed by randomly flipping the image horizontally. Finally, we crop $128\times128$ pixel region from the image and converted to grayscale. During testing, we consider the center $128\times128$ pixel crop from the image and converted it to grayscale. 
The trained gender-prediction models are evaluated on $19,962$ and $12,000$ test samples from CelebA and VISOB, respectively. The age-group classification model is evaluated on $3,000$ test samples samples from Adience dataset. The test accuracy values of the gender and age-prediction models trained using supervised and self-ensemble based learning using various combination of labeled and unlabeled data are shown in Table~\ref{table2}~\ref{table3}, and~\ref{tab:age_group}.

\begin{table}[]
\label{tab:gender_celeba}
\caption{Gender prediction performance comparison of the deep learning model trained using supervised and our proposed self-ensemble method (supervised\% /semi-supervised\%) on face images from \textbf{CelebA} dataset.}
\begin{center}
\begin{tabular}{c|c|ccc}
\hline
\textbf{\begin{tabular}[c]{@{}c@{}}Labeled \\ Samples\end{tabular}} &
\textbf{\begin{tabular}[c]{@{}c@{}}Unlabeled \\ Samples\end{tabular}} &\textbf{ MCR($\%$)} & \textbf{FCR($\%$)} & \textbf{ACC(\%)} \\ \hline\hline
\textbf{All samples}                                                 & \textbf{0} & 96.8/-                                                            & 98.51/-                                                             & 97.91/-                                                            \\
\textbf{100}                                                         & \textbf{199900} & 0/75.49                                                           & 100/89.67                                                           & 64.8/\textbf{84.68}                                                \\
\textbf{200}                                                         & \textbf{199800} & 3.67/90.45                                                        & 98.86/92.45                                                         & 65.35/\textbf{91.74}                                               \\
\textbf{500}                                                         & \textbf{199500} & 49.42/92.55                                                       & 95.44/94.62                                                         & 79.24/\textbf{93.89}                                               \\
\textbf{1000}                                                        & \textbf{199000} & 84.87/94.5                                                        & 90.81/94.44                                                         & 88.72/\textbf{94.46}                                               \\ \hline
\end{tabular}
\end{center}
\label{table2}
\end{table}

\begin{table}[]
\label{tab:gender_visob}
\caption{Gender prediction performance comparison of the deep learning model trained using supervised and our proposed self-ensemble based method (supervised\%/semi-supervised\%)  on ocular region images from \textbf{VISOB} dataset.}
\begin{center}
\begin{tabular}{c|c|ccc}
\hline
\textbf{\begin{tabular}[c]{@{}c@{}}Labeled \\ Samples\end{tabular}} & \textbf{\begin{tabular}[c]{@{}c@{}}Unlabeled \\ Samples\end{tabular}} & \textbf{ MCR($\%$)} & \textbf{FCR($\%$)} & \textbf{ACC(\%)}  \\ \hline\hline
\textbf{All samples}                                                 & \textbf{0} &  92.16/-                                                           & 75.94/-                                                             & 84.65/-                                                           \\
\textbf{100}                                                         & \textbf{108335} & 100.00/80.01                                                      & 0.00/57.32                                                          & 53.70/\textbf{69.51}                                              \\
\textbf{200}                                                         & \textbf{108235} & 92.47/76.42                                                       & 9.11/71.33                                                          & 53.87/\textbf{74.07}                                              \\
\textbf{500}                                                         & \textbf{107935} & 84.79/85.19                                                       & 47.71/66.41                                                         & 67.63/\textbf{76.49}                                              \\
\textbf{1000}                                                        & \textbf{107435} & 88.75/83.40                                                       & 53.71/78.22                                                         & 72.53/\textbf{81.00} \\
\hline
\end{tabular}
\end{center}
\label{table3}
\end{table}

\begin{table*}[]
\caption{Exact and 1-off accuracy of age-group prediction for the  deep learning model trained with supervised (baseline) and our proposed self-ensemble method for 5-fold cross-validation. The overall performance is shown as mean $\pm$ standard deviation on \textbf{Adience} dataset.}
\begin{center}

\begin{tabular}{c|cccc}
\hline
\multirow{2}{*}{\textbf{\begin{tabular}[c]{@{}c@{}}Cross \\ Validation\end{tabular}}} & \multicolumn{2}{c}{\textbf{Exact Acc(\%)}}     & \multicolumn{2}{c}{\textbf{1-OFF Acc(\%)}} \\ \cline{2-5} 
                                                                                      & \textbf{baseline} & \textbf{proposed}      & \textbf{baseline}  & \textbf{proposed} \\ \hline\hline
1                                                                                     & 49.55             & 54.39                  & 84.6               & 87.3              \\
2                                                                                     & 48.28             & 55.1                   & 87.67              & 90.14             \\
3                                                                                     & 54.46             & 52.87                  & 88.68              & 88.71             \\
4                                                                                     & 48.46             & 51.64                  & 86.41              & 85.59             \\
5                                                                                     & 57.42             & 63.76                  & 90.06              & 91.88             \\ \hline\hline
\textbf{Overall}                                                                      & 51.63 $\pm$ 3.66     & \textbf{55.55 $\pm$ 4.28} & 87.48 $\pm$ 1.87    & \textbf{88.72 $\pm$ 2.18}   \\ \hline\hline
\textbf{From~\cite{levi2015age}}                                                                  & 50.7 $\pm$ 5.1       &                        & 84.7 $\pm$ 2.2        &                   \\ \hline
\end{tabular}
\end{center}
\label{tab:age_group}
\end{table*}

\begin{table}[]
\label{tab:gender_celeba}
\caption{Gender prediction performance comparison of our proposed self-ensemble method (supervised\% /semi-supervised\%) with FixMatch~\cite{sohn2020fixmatch} which was implemented  on face images from \textbf{CelebA} dataset.}
\begin{center}
\begin{tabular}{c|c|cc}
\hline
\textbf{\begin{tabular}[c]{@{}c@{}}Labeled \\ Samples\end{tabular}} &
\textbf{\begin{tabular}[c]{@{}c@{}}Unlabeled \\ Samples\end{tabular}} &\textbf{ACC(\%)}\textbf{FixMatch}&\textbf{ACC(\%)}\textbf{Proposed} \\ \hline\hline
\textbf{All samples}    &                                              \textbf{0} &       97.8/-                                                                                                                & 97.91/-                                                            \\
\textbf{100}                                                         & \textbf{199900}& -/\textbf{91.9}                                                           & 64.8/\textbf{84.68}                                                \\
\textbf{200}                                                         & \textbf{199800}                                                         &     -/\textbf{92.85}&65.35/\textbf{91.74}                                               \\
\textbf{500}                                                         & \textbf{199500}                                                        & -/\textbf{93.25}&79.24/\textbf{93.89}                                               \\
\textbf{1000}                                                        & \textbf{199000}                                                         & -/\textbf{93.65}&88.72/\textbf{94.46}                                               \\ \hline
\end{tabular}
\end{center}
\label{table4}
\end{table}

\section{Results}
\label{sec:results}

Table~\ref{table2} and Table~\ref{table3} show gender prediction accuracy for supervised and self-ensemble model for face images from CelebA dataset and ocular images from VISOB dataset, respectively. The number of labeled and unlabeled samples and the obtained accuracy values using supervised as well as self-ensemble based models are also shown. It can be seen that using the low number of labeled samples, \textbf{$100$}  and \textbf{$200$} samples, the supervised model performance is dropped with about $30\%$ accuracy in predicting the gender. Whereas, with the proposed self-ensemble model better capability at prediction the gender while at least \textbf{$15\%$} improvement in total prediction accuracy (ACC\%) could be obtained. 

It can also be seen that the accuracy of the proposed self-ensemble learning with only \textbf{$1000$} labeled samples obtained \textbf{$81\%$} accuracy on VISOB dataset and \textbf{$94.46\%$} on CelebA in gender prediction, which is very close to the model trained in a supervised manner with all the training dataset as labeled samples with {$84.65\%$ and $97.91\%$} accuracy for VISOB and CelebA, respectively. \textit{The experimental results demonstrate that the self-ensemble model for gender prediction trained using both limited labeled data (as low as 100 samples) and unlabeled data (about 108K-199K) obtains performance equivalent to supervised model trained on all 108K-199K labeled samples.}

Table~\ref{tab:age_group} show the age-group prediction on Adience dataset with the deep learning model trained using supervised learning (with about 11K samples) as the baseline. The comparative analysis is done with the proposed self-ensemble model for 5-fold cross-validation. It can be seen that the supervised deep learning model trained on 11K samples can obtain exact accuracy of \textbf{$51.63\% \pm 3.66$} which already is \textbf{$1\%$} higher than the model proposed in \cite{levi2015age} with exact accuracy of \textbf{$50.7\% \pm 5.1$}. On top of that, by using our proposed self-ensemble learning (using 11K samples as labeled and unlabeled data equally divided),  better exact accuracy of \textbf{$55.55\% \pm 4.28$} which is \textbf{$3.92\%$} more than the baseline, could be obtained. 

Table~\ref{table4} show the gender prediction performance comparison of the proposed self-ensemble method with the FixMatch~\cite{sohn2020fixmatch}. Both the models were evaluated on face images from the CelebA dataset.

\textbf{The experimental results on gender and age prediction from face and ocular images suggest the efficacy of harnessing unlabeled data towards improving generalization accuracy of the deep learning models.}

\section{Conclusion and Future Work}
\label{sec:conclusion}
The generalization ability of the deep learning models is proportional to the use of large scale and representative datasets for model training. However, labeling the large scale dataset require human operator and is very expensive and time consuming. In this paper, we proposed a self-ensemble based semi supervised learning model which could be trained using labeled data in a supervised manner and can consecutively harness the unlabeled data in an unsupervised fashion using multi-objective loss function. A case study on gender and age-group prediction using face and ocular images from CelebA, Adience and VISOB datasets suggest increase in the generalization ability of the proposed deep learning model when trained with small amount of labeled data (as less as 100 samples) along with large unlabeled dataset. The deep learning models with the ability to harness unlabeled data also have applications in few shot learning~\cite{yao2021crossdomain}, bias mitigation towards unrepresentative sub-population in the dataset~\cite{krishnan2020understanding}, and model adaptation to the dynamic and temporal variations of the operational data~\cite{chen2020action}. 
As a part of future work, comparative analysis of the proposed self-ensemble method, implemented using different backbone architectures such as ResNet and MobileNet, with the other semi supervised learning methods like MixMatch~\cite{berthelot2019mixmatch}, Noisy Student~\cite{xie2020selftraining} and FixMatch~\cite{sohn2020fixmatch} will be performed.

\balance
\bibliographystyle{IEEEtran}
\bibliography{IEEEabrv,biblio_traps_dynamics}

\end{document}